\documentclass[sigconf]{acmart}
\renewcommand\footnotetextcopyrightpermission[1]{} 
\renewcommand\footnotetextcopyrightpermission[1]{}

\AtBeginDocument{%
  \providecommand\BibTeX{{%
    \normalfont B\kern-0.5em{\scshape i\kern-0.25em b}\kern-0.8em\TeX}}}

\setcopyright{acmcopyright}
\copyrightyear{2018}
\acmYear{2018}
\acmDOI{10.1145/1122445.1122456}

\acmConference{*}{Arxiv Version}{For Review}
  
\acmBooktitle{Woodstock '18: ACM Symposium on Neural Gaze Detection,
  June 03--05, 2018, Woodstock, NY}
\acmPrice{15.00}
\acmISBN{978-1-4503-XXXX-X/18/06}

\usepackage{booktabs, colortbl}
\usepackage{multirow}

\newcommand{\best}[1]{{\color{blue}{\textbf{#1}}}} 
\newcommand{\second}[1]{{\color{black}{\textbf{#1}}}}

\usepackage{xspace}
\newcommand{\oursframe}{UNIAA\xspace}
\newcommand{\oursmodel}{UNIAA-LLaVA\xspace}
\newcommand{\oursbench}{UNIAA-Bench\xspace}
\usepackage{graphicx}
\usepackage{caption}

\def\logo{\makebox[0pt][l]{\hspace{0pt}\raisebox{-0.3ex}{\includegraphics[height=19pt]{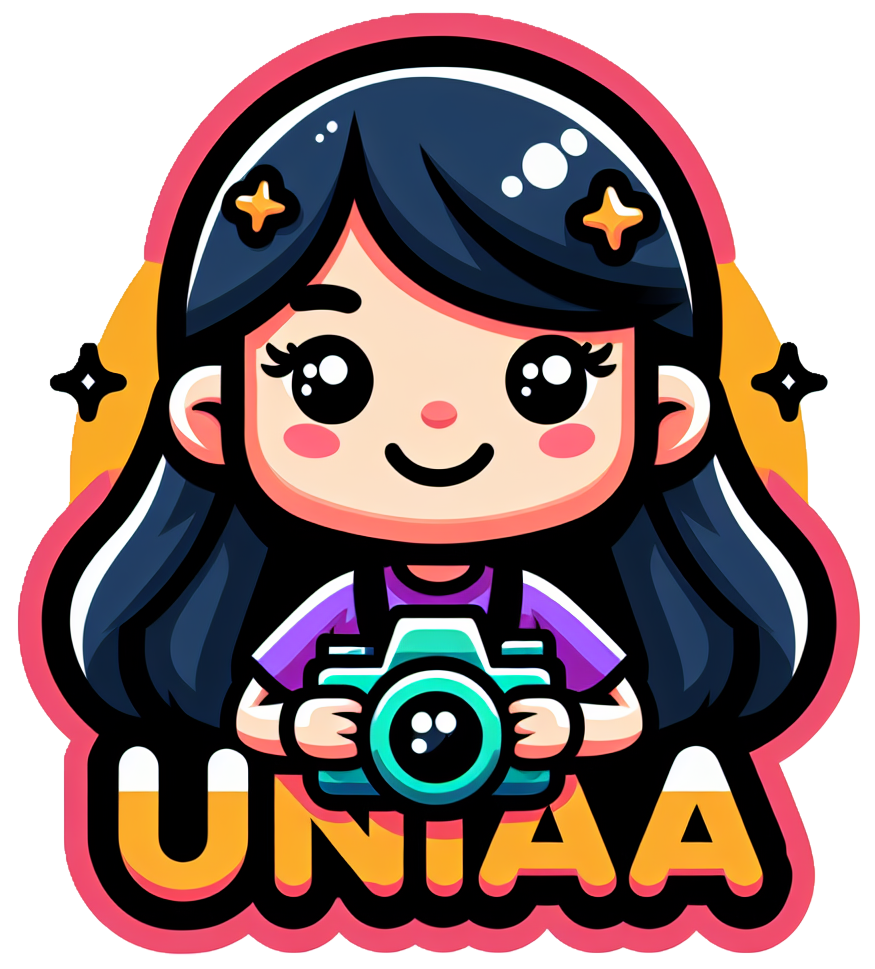}}}}

\begin{document}

\title[A Unified Multi-modal Image Aesthetic Assessment Baseline and Benchmark]{\logo\ \ \ \ \oursframe: A Unified Multi-modal Image Aesthetic Assessment Baseline and Benchmark}

\author{Zhaokun Zhou}
\authornote{This work was done during internship at Kuaishou Technology. Company mentor is Qiulin Wang.}
\affiliation{
  \institution{School of Electronic and Computer\\ Engineering, Peking University}
  \city{Shenzhen}
  \country{China}
}

\author{Qiulin Wang}
\affiliation{
\institution{Kuaishou Technology}
  \city{Beijing}
  \country{China}
}

\author{Bin Lin}
\affiliation{
  \institution{School of Electronic and Computer\\ Engineering, Peking University}
  \city{Shenzhen}
  \country{China}
}

\author{Yiwei Su}
\affiliation{
\institution{Kuaishou Technology}
  \city{Beijing}
  \country{China}
}

\author{Rui Chen}
\affiliation{
\institution{Kuaishou Technology}
  \city{Beijing}
  \country{China}
}
\author{Xin Tao}
\authornote{Corresponding author. taoxin@kuaishou.com}
\affiliation{
\institution{Kuaishou Technology}
  \city{Beijing}
  \country{China}
}
\author{Amin Zheng}
\affiliation{
\institution{Kuaishou Technology}
  \city{Beijing}
  \country{China}
}

\author{Li Yuan}
\affiliation{
  \institution{School of Electronic and Computer\\ Engineering, Peking University}
  \city{Shenzhen}
  \country{China}
}

\author{Pengfei Wan}
\affiliation{
\institution{Kuaishou Technology}
  \city{Beijing}
  \country{China}
}

\author{Di Zhang}
\affiliation{
\institution{Kuaishou Technology}
  \city{Beijing}
  \country{China}
}

\begin{abstract}
As an alternative to expensive expert evaluation, Image Aesthetic Assessment (IAA) stands out as a crucial task in computer vision.
However, traditional IAA methods are typically constrained to a single dataset or task, restricting the universality and broader application.
In this work, to better align with human aesthetics, we propose a \textbf{Un}ified Multi-modal \textbf{I}mage \textbf{A}esthetic \textbf{A}ssessment (\textbf{\oursframe}) framework, including a Multi-modal Large Language Model (MLLM) named \textbf{\oursmodel} and a comprehensive benchmark named \textbf{\oursbench}.
We choose MLLM with both visual perception and language ability for IAA and establish a low-cost paradigm for transforming the existing datasets into unified and high-quality visual instruction tuning data, from which the \oursmodel is trained. To further evaluate the IAA capability of MLLMs, we construct \oursbench, which consists of three aesthetic levels: Perception, Description, and Assessment.
Extensive experiments validate the effectiveness and rationality of \oursframe.
\oursmodel achieves competitive performance on all levels of \oursbench, compared with existing MLLMs.
Specifically, our model performs better than GPT-4V in aesthetic perception and even approaches the junior-level human.
We find MLLMs have great potential in IAA, yet there remains plenty of room for further improvement.
The \oursmodel and \oursbench will be released.

\end{abstract}



\maketitle

\section{Introduction}
Image Aesthetics Assessment (IAA) in computer vision involves the evaluation of the photography techniques and artistic approaches of images \cite{deng2017image}, to categorize them into various aesthetic levels \cite{hou2022distilling,niu2022comment}.
Serving as an alternative to costly expert aesthetic evaluation, IAA has a wide range of applications in image retrieval \cite{liu2021infrared}, album curation \cite{guo2017multigranular}, smart photography \cite{rawat2016clicksmart}, and image editing\cite{chai2020roundness}, which is still a challenging task \cite{yang2019comprehensive}.
Various datasets and models for IAA have been proposed, achieving significant progress.
IAA datasets offer images with aesthetic scores \cite{murray2012ava, kong2016photo,Yang_2022_CVPR} and other annotations of aesthetic attributes \cite{kong2016photo, he_2022_ijcai, Yang_2022_CVPR, murray2012ava}, as well as overall description \cite{jin2019aesthetic}. 
Many studies have achieved promising results by employing data-driven approaches to directly predict the aesthetic mean opinion scores (MOS) \cite{he_2022_ijcai,xiong2023image,he2023eat}, including convolutional neural networks \cite{talebi2018nima, Chen_2020_CVPR, he_2022_ijcai, Hosu_2019_CVPR}, and transformers \cite{Ke_2021_ICCV,xia2022vision,he2023eat}.
There are also works that use aesthetic attributes and image content to assist in predicting scores \cite{kong2016photo,Yang_2022_CVPR,he2023thinking}.
VILA \cite{ke2023vila} and RPCD \cite{vera2022understanding} have demonstrated the potential of learning image aesthetics from photo comments.
\begin{figure*}
\centering
\includegraphics[scale=0.7]{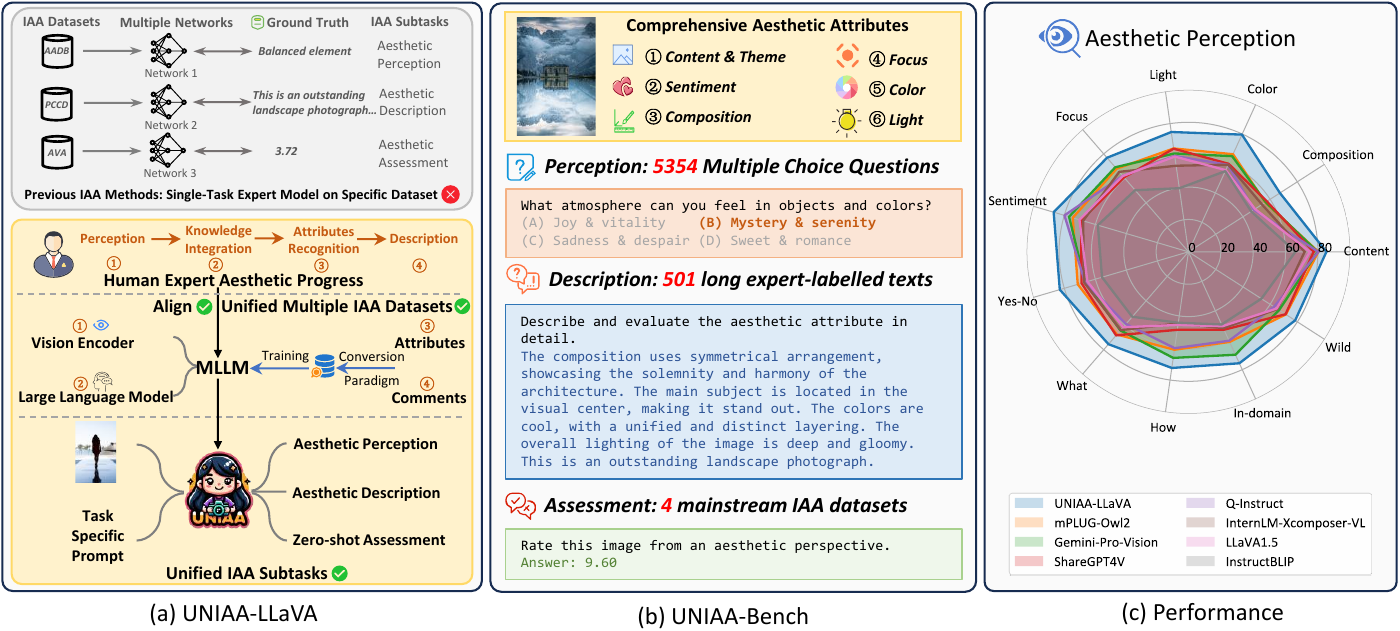}
\caption{The Unified Multi-modal Image Aesthetic Assessment Framework, containing a baseline (a) and a benchmark  (b). The aesthetic perception performance of \oursmodel and other MLLMs is shown in (c).}
\label{fig:teaser}
\end{figure*}

Psychological studies reveal that the human experience of aesthetics involves a series of information processing stages, encompassing perception, integration of implicit memory, explicit classification of content and style, and evaluation \cite{leder2004model,chatterjee2003prospects}.
However, current IAA methods are not well-aligned with the human aesthetic process.
Regarding IAA datasets, most of them only cover a partial level of the human aesthetic evaluation process and lack standardization in their formats, which makes it difficult to train them under a uniform architecture.
For example, AVA contains both the MOS of images as well as rich comments, photography style labels, and content tags. 
Aesthetic annotations are represented in different formats consisting of numeric, language, and category/attribute labels.
Regarding IAA methods, they only utilize a part of aesthetic information and are predominantly focused on a single aesthetic task.
Scoring models struggle to recognize the aesthetic attributes of images, while aesthetic description models can not score photos.

We need a unified IAA framework that can \textbf{align well with human aesthetic processes}, \textbf{integrate different sources and formats of aesthetic data}, and \textbf{achieve good results in multiple aesthetic subtasks}.
Specifically, we argue that in the process of IAA-model learning, it is necessary to align with the process of human aesthetics, rather than simply relying on implicit mapping through learning MOS or partial attributes \cite{vera2022understanding,ke2023vila}.
The model necessitates a visual understanding and harbors inherent reservoirs of aesthetic knowledge. 
It needs to learn explicit categorization of the aesthetic attributes and integrate the aesthetic information to provide detailed descriptions and overall scores.
MLLM \cite{gpt4v,flamingo,blip2,palme,minigpt4,llava,kosmos,videochat} is primarily composed of a vision encoder and an LLM, which is capable of visual perception and has substantial knowledge capacity \cite{schaeffer2023emergent}, making it a natural fit for human-aligned and unified IAA.

We propose \textbf{\oursframe}, as shown in Figure \ref{fig:teaser}, which to the best of our knowledge, introduces MLLM to IAA for the first time, including an IAA-MLLM baseline and a comprehensive aesthetic benchmark.
The baseline is capable of unifying aesthetic perception, description, and assessment tasks, while the benchmark comprehensively evaluates the aesthetic capabilities of MLLM from these three aspects, which strives to align the aesthetic evaluation process with that of human experts.
For the baseline model, we adopt the well-established LLaVA \cite{liu2023visual,liu2023improved} architecture and further finetune it with aesthetic visual instruction tuning data to obtain our \textbf{\oursmodel} (Figure \ref{fig:teaser}(a)).
Based on the integration of visually aesthetic perception by the vision encoder and memory of aesthetic knowledge through LLM, \oursmodel demonstrates a strong ability to learn and align with human aesthetics. 
However, due to the subjective and complex nature of aesthetics \cite{Yang_2022_CVPR}, annotating a large amount of aesthetic visual instruction tuning data is very expensive.
Therefore, we establish a low-cost IAA Dataset Conversion Paradigm (IDCP), aiming to convert aesthetic annotations such as aesthetic attributes and comments from off-the-shelf IAA datasets into a MLLM visual-instruction-tuning format.

To evaluate the aesthetic ability of the proposed \oursmodel and other existing MLLMs, we construct a comprehensive and professional \textbf{\oursbench} (Figure \ref{fig:teaser}(b)).
To ensure diversity, we select test-set images from the IAA datasets as the \textbf{in-domain} split and collect images from the internet as the \textbf{in-the-wild} split. The in-the-wild split does not overlap with the IAA datasets that are used to train the \oursmodel, which can further prove its generalization capability on various aesthetic tasks.
In accordance with the image-quality Q-bench \cite{wu2023qbench} and human aesthetic \cite{leder2004model}, we evaluate the aesthetic ability of MLLMs from three aspects, including \textbf{Perception}, \textbf{Description}, and \textbf{Assessment}.
\textbf{For Aesthetic Perception}, we construct the \oursframe-QA dataset, which covers six general aesthetic attributes \cite{Yang_2022_CVPR, li2023theme} through its questions: \textbf{content and theme, composition, color, light, focus and sentiments}.
\textbf{For Aesthetic Description}, we construct the \oursframe-Describe dataset to assess the ability of MLLM on detailed and professional aesthetic comments and suggestions.
\textbf{For Aesthetic Assessment}, we test the zero-shot scoring ability of MLLMs through output token logits.
Our contributions can be summed up into four points:

$\bullet$   We establish an IAA framework that includes an IAA-Baseline
and a comprehensive IAA-Bench, namely \textbf{\oursframe}. Both the learning process and evaluation process of \oursframe align with the visual
aesthetic process of human experts.

$\bullet$ We develop an IAA Datasets Conversion Paradigm (IDCP) to
convert existing aesthetic datasets into a format suitable for MLLMs
fine-tuning. Using converted data, we obtain the \textbf{\oursmodel}, which unifies IAA subtasks with language.

$\bullet$ We are constructing a comprehensive aesthetic evaluation benchmark called \textbf{\oursbench}, which systematically evaluates the aesthetic abilities of MLLMs from three overall dimensions: Aesthetic Perception, Aesthetic Description, and Aesthetic Assessment.

$\bullet$ We demonstrate the effectiveness of \oursmodel and IDCP, the authority of \oursbench, and the feasibility of MLLM in accomplishing aesthetic tasks through experiments.

\section{Related Work}
\subsection{Image Aesthestic Assessment}\label{rw_IAA}
Image Aesthetics Assessment has three main tasks: direct scoring of the aesthetic quality of images \cite{murray2012ava,karayev2013recognizing,tang2013content,Ren_2017_ICCV,lee2018photographic,Yang_2022_CVPR,he_2022_ijcai}, recognition of aesthetic attributes \cite{celona2022composition,jin2019aesthetic}, and aesthetic description \cite{zhou2016joint,wang2019neural}.
A multitude of approaches, encompassing network architecture design \cite{tang2013content,he2023eat}, loss function design \cite{kong2016photo,talebi2018nima}, utilization of the CLIP \cite{clip} pre-training model \cite{ke2023vila,xiong2023image}, and more, have been proposed, demonstrating strong performance on their respective tasks and datasets.
However, the majority of IAA models focus on accomplishing one or two subtasks. 
There is a lack of a comprehensive and versatile IAA model for holistic image aesthetic evaluation.

Numerous datasets emerge to support the advancement of IAA.
Murray et al. \cite{murray2012ava} pioneered the AVA dataset, which has gained significant popularity. Kong et al. \cite{kong2016photo} introduced the Aesthetics and Attributes Database (AADB), encompassing subjective aesthetic scores and attributes 
Addressing the need for generating captions based on photo aesthetics, Yu et al. \cite{chang2017aesthetic} developed the Photo Critique Captioning Dataset (PCCD). 
TAD66K \cite{he_2022_ijcai} is an extensive dataset encompassing 47 subjects.
 The Personalized Image Aesthetics Database with Rich Attributes (PARA) \cite{Yang_2022_CVPR} includes nine image-oriented objective attributes and four human-oriented subjective attributes.
ICAA17K \cite{he2023thinking} comprises 17K images capturing 30 popular color combinations, providing a valuable resource for color-oriented analysis.
While these datasets were previously handled separately, the shared and correlated nature of aesthetic knowledge provides the potential for integrating learning.

However, the current IAA methods still have a significant gap from a general-purpose image aesthetic evaluation, mainly due to the following two issues.
Firstly, relying on human-labeled aesthetic ratings, i.e. MOS, for directly supervised learning may be sub-optimal, as it lacks of interpretability \cite{kang:hal-02934292} and may lead to overfitting to label noise. Moreover, current methods are still constrained by the need to train different models for individual datasets. 
Tasks such as aesthetic perception, aesthetic description, and aesthetic assessment also require training in separate models \cite{yang2019comprehensive}. 
Current IAA methods lack comprehensive aesthetic information learning and universal applicability across different datasets and tasks.
These limitations hinder the practical applicability and universality of IAA.
Therefore, IAA urgently requires a unified and versatile approach to enhance its practical value.

\subsection{Multi-modality Large Language Models}
Large language models (LLMs) such as GPT-4~\cite{openai2023gpt4}, T5~\cite{flant5}, and LLaMA~\cite{llama}, have demonstrated a remarkable language proficiency in general knowledge. 
By incorporating vision encoders~\cite{clip} and additional adaptive modules into LLMs, Multi-modal Large Language Models (MLLMs)~\cite{otter,llamaadapterv2,llava,iblip,xcomposer}  have driven significant progress in tackling various multi-modal tasks concerning high-level vision, such as image captioning~\cite{cococaps,nocaps,flickrcaps}, visual question answering (VQA)~\cite{cocovqa,okvqa,AOKVQA}, and other language-related abilities~\cite{mmbench,mme,seedbench}.

Although MLLMs \cite{gpt4v,flamingo,blip2,palme,minigpt4,llava,kosmos,videochat, lin2023video,lin2024moe,zhu2023languagebind, jin2023chatunivi} play a crucial role in advancing vision-language learning, they exhibit subpar performance in specific tasks and vertical applications \cite{seedbench,mme,wu2023q,li2023llava}. 
The emergence of fine-tuning with task-specific instructional knowledge has led to a surge of MLLMs tailored for specific tasks, such as Q-instruct \cite{wu2023qinstruct}, LLaVA-Med \cite{li2023llava}, and Drive-GPT4 \cite{xu2023drivegpt4}.

\noindent\textbf{\textit{Why choose MLLM for IAA?}}
There are two reasons for adapting MLLMs to IAA tasks.
\textbf{(1)} MLLMs are capable of aesthetic visual perception and knowledge integration. 
\textbf{(2)} The fine-tuning data and output are both mediated by language, which facilitates the natural integration of aesthetic visual instruction tuning and enables seamless completion of IAA subtasks.
These training data and subtasks are highly associated with the human aesthetic process \cite{leder2004model}.
Our goal is to create a unified aesthetic foundational model that can integrate various aesthetic information and establish general capabilities across various tasks while robustly responding to open-ended human queries.

\begin{figure*}
\centering
\includegraphics[scale=0.76]{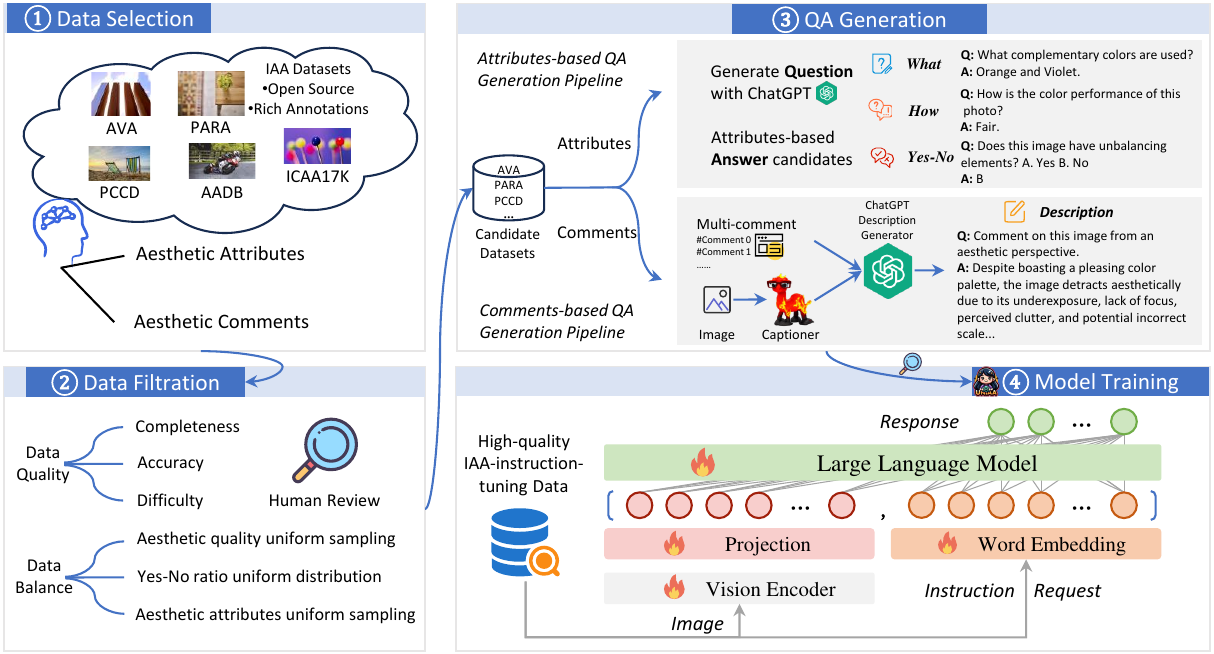}
\caption{\textbf{The IAA Datasets Conversion Paradigm for \oursmodel.}}
\label{Fig:uniaa_llava_overview}
\end{figure*}

\section{\oursmodel}
In the following two sections, we present the details of our proposed aesthetic evaluation framework, \oursframe, which aims to unify the aesthetic information perception, description, and assessment tasks.
As a starting point, we integrate LLaVA, a versatile multimodal conversational model~\cite{liu2023visual}, as the initial model. 
Subsequently, we consistently refine and fine-tune the model to specifically cater to the demands of the aesthetic domain.
There is a shortage of multi-modal image aesthetic datasets for training an instruction-following assistant. 
To bridge this gap, we construct the initial dataset of its kind, derived from extensively available IAA datasets.
As a fundamental part of constructing the instruction fine-tuning dataset, we first introduce a low-cost paradigm for converting the instruction fine-tuning data. 
Unlike the need for manual annotation collection, we utilize the existing IAA dataset and transform it into corresponding suitable IAA instruction fine-tuning data based on the aesthetic dimensions emphasized during its labeling and annotation settings. 
Then, further using the transformed data, we obtain the first MLLM of IAA via instruction fine-tuning on the LLaVA.

\subsection{IAA Datasets Conversion Paradigm}
Inspired by the construction of classic datasets of instruction tuning~\cite{mme, mmbench, wu2023qinstruct,chen2023sharegpt4v}, our IAA Data Conversion Paradigm (IDCP) consists of three steps: \textbf{data selection}, \textbf{data filtering}, and \textbf{question-answer generation}, as illustrated in Figure \ref{Fig:uniaa_llava_overview}.
The converted data will be used for training \oursmodel. 
We hope that our IAA data transformation pipeline can provide inspiration and guidance for building new IAA instruction tuning datasets as well as other domain-specific task datasets.

\subsubsection{\textbf{\textit{{Data Selection.}}}\\}
To maximize the diversity and comprehensiveness of \oursframe, we first select five open-source mainstream IAA datasets, AVA, AADB, ICAA17K, PARA, and PCCD, detailed in Appendix \ref{appendix:drd}. 
These datasets have aesthetic annotations across different dimensions.
We select specific aesthetic information.
The annotation content of existing IAA datasets can be primarily categorized into the following three types, \textbf{aesthetic attributes}, \textbf{aesthetic comments}, and \textbf{aesthetic scores}. 
We choose the first two categories.

\textbf{Aesthetic Attributes} are annotated information about photographs on specific objective aesthetic dimensions.
Its exceptional suitability for conversion into visual question answering empowers the model with the ability to recognize and evaluate the aesthetic attributes of images using concise words.
The data sources for aesthetic attributes include the photography styles annotated in AVA, the influence of aesthetic attributes on overall image aesthetics in AADB, the color schemes in the ICAA17K dataset, and the quality classification of aesthetic attributes in PARA.

\textbf{Aesthetic Comments} are evaluations and descriptions of aesthetic features, aiming to explore and convey the aesthetic ambiance and emotional experiences that images invoke. 
Descriptions of aesthetic comments can be made from various perspectives, such as color, composition, form of expression, meaning, emotions, etc. 
These comments can help models better understand and appreciate images. 
The data sources for aesthetic comments include the overall comment on a photo in AVA and the dimensional comment in PCCD including the use of camera, photo subject, depth of field, focus, color brightness, composition, and overall impression.

 \noindent\textbf{\textit{Why not use Aesthetic Scores?}}
Aesthetic Scores are highly subjective \cite{Yang_2022_CVPR}.
Due to the differences in scoring rules and aesthetic differences among annotation experts, it is difficult to convert scores from different IAA datasets into unified evaluation scores.
For example, a photo with a high score in one dataset may not have better aesthetic quality than a photo with a medium score in another dataset.
Instead, we use attributes and descriptions to enable the \oursmodel to have good zero-shot scoring capability, detailed in Section \ref{raa}. 
As a supplementary experiment, based on the vision encoder of \oursmodel, we further conduct supervised training on specific datasets and achieve the SOTA performance, demonstrating the generalization of our learning framework.

\subsubsection{\textbf{\textit{{Data Filtration.}}}\\}
\textbf{Data Quality.} We ensure the quality of the training data from the source by conducting manual checks on the selected data for both completeness and accuracy.
We filter out empty or incomplete annotations and remove incorrectly labeled aesthetic dimensions.
We check the difficulty of aesthetic information by removing data that is too simple or too difficult.
We discard the "Tags" of AVA, which contain simple categories like "Water" and "Street" due to their low relevance to aesthetics.
For some attributes in AADB, such as whether colors are harmonious, we filter out samples that are difficult for experts to judge.

\textbf{Data Balance.}
To introduce diverse visual aesthetic appearances, we uniformly sampled images from each aesthetic quality range.
We balance the number of aesthetic attributes as well as the number of options within each attribute.
For yes or no questions, we ensure the balance between "yes" and "no" responses.

\subsubsection{\textbf{\textit{{QA Generation.}}}\\}
\textbf{Attributes-based Pipeline.} 
According to aesthetic attribute annotation, aesthetic experts first manually write down several elemental questions, and then let ChatGPT expand them into various forms while ensuring the same meaning, detailed in Appendix \ref{appendix-pqgaa}.
To imitate diverse query formats from human beings \cite{wu2023qbench}, we organize three question types, Yes-or-No, What, and How, which correspond to different types of annotations.
For binary attribute questions, such as whether elements in a photo are balanced, "yes/no" is used.
Specifically, for questions about the types of attributes, such as the primary color usage of a picture, "what" is used. 
For quality classification questions, such as the quality of focus, "how" is used. 

\begin{figure*}
\centering
\includegraphics[scale=0.66]{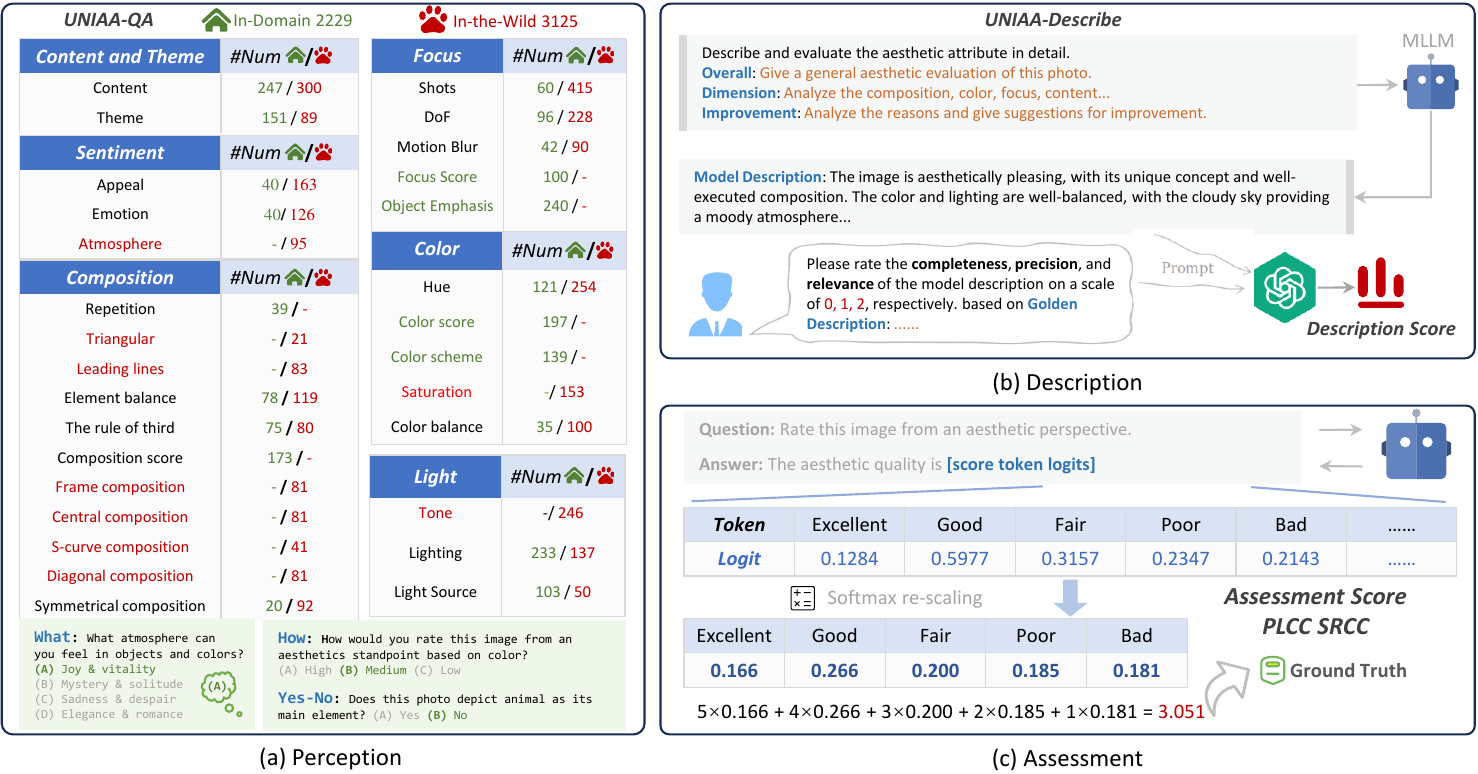}
\caption{\textbf{The \oursbench overview.} (a) \oursframe-QA contains 5354 Image-Question-Answer samples and (b) \oursframe-Describe contains 501 Image-Description samples. (c) For open-source MLLMs, Logits can be extracted to calculate the score.}
\label{fig:uniaa-bench}
\end{figure*}
\setlength{\abovecaptionskip}{2pt} 

\textbf{Comments-based Pipeline.} 
The quality and style of aesthetic comments in different datasets can vary. 
For instance, the comments in the AVA \cite{murray2012ava} are more colloquial, whereas the PCCD \cite{chang2017aesthetic} contains professional comments from experienced photographers. 
It is necessary to use ChatGPT \cite{chatgpt,gpt4} to clean the raw data, detailed in Appendix \ref{app:pdg}. 
The transformation process of aesthetic comments mainly consists of two steps: comments rewriting and question matching.  
We utilize GPT and LLaVA to assist in rewriting the comments. 
LLaVA offers the image content and details. 
GPT is prompted to filter out the drawbacks of aesthetic comments from different datasets, such as the colloquial content in AVA that is not relevant to aesthetics and the excessively long comments in PCCD.

Regarding question matching, we use the length and dimension of cleaned content as key points. 
We ask the MLLM questions to evaluate the image's overall aesthetic and specific dimensions, such as composition, lighting, and impression.
If the comments include explanations for aesthetic patterns and suggestions for aesthetic improvements, the matching questions will also include phrases such as "analyze the causes of this characteristic" and "please provide some aesthetic improvement suggestions."
If the comments exceed a certain length, such as 100 words, the corresponding question would be to provide a ``detailed'' evaluation of the image.

\subsubsection{\textbf{\textit{{IDCP Advantages Discussion.}}}\\}
\noindent As a general MLLM aesthetic data conversion paradigm, IDCP has the following three advantages.
Firstly, it is \textbf{low cost}. IDCP integrates publicly available datasets, without the need for additional manpower to collect and annotate data.
Secondly, it has \textbf{uniformity}. Through conversion, aesthetic comments and aesthetic attributes can be trained in a unified format, enhancing the overall aesthetic capabilities of the model.
Thirdly, it has \textbf{scalability}. In addition to the five datasets we select, it can also be used to incorporate newly emerged aesthetic datasets, and more data will further enhance the aesthetic capabilities of the model.

\subsection{Image Aesthetic Visual Instruction Tuning}
In this section, we discuss the training approaches used to fine-tune MLLMs with Image Aesthetic visual instructions.
The training process for open-source MLLMs ~\cite{iblip,otter,xcomposer} consists of two stages. 
The first stage involves aligning the representation space of the visual backbone and LLM with million-scale data ~\cite{laion400m, cc3m}. 
The second stage includes fine-tuning the MLLM with a combination of human-labeled datasets~\cite{refcoco,cocovqa,cococaps,okvqa} for visual instruction tuning.
Considering the scale and computational cost of the IDCP dataset, we utilize the MLLM fine-tuned after the second stage for visual instruction tuning to perform IAA learning.
IAA tasks require a finer level of visual perception. 
We open the parameter training of the vision encoder during training.

\section{\oursbench}
It is uncertain how well the \oursmodel performs in image aesthetics. 
The lack of a dedicated benchmark for assessing the effectiveness of MLLMs in aesthetic evaluation presents an apparent obstacle, which may hinder the progress of advanced MLLMs capable of aesthetics.
To solve this problem, we propose \oursbench, an expert benchmark designed to comprehensively evaluate the aesthetic capabilities of MLLMs.
To ensure the diversity of the benchmark~\cite{liu2023visual}, we select 
images from the internet and other datasets that have never been seen by \oursmodel, serving as \textbf{In-the-Wild} resources.
IAA-Bench also contains test-set images from the IAA datasets as \textbf{In Domain} resources, whose training sets were previously used during the training of our baseline. 
Inspired by human aesthetic process~\cite{leder2004model} and image-quality bench ~\cite{wu2023qbench}, we suggest a collection of integrative standards to measure the aesthetic skills of MLLMs from three perspectives: \textbf{Aesthetic Perception}, \textbf{Aesthetic Description}, and \textbf{Aesthetic Assessment}.

\subsection{Aesthetic Perception}
We construct the \oursframe-QA to investigate whether MLLMs can accurately respond to simple natural language queries related to aesthetic attributes, which is also an ability of human experts to recognize explicit aesthetic information ~\cite{leder2004model}.
\oursframe-QA contains 5,354 images, as listed in Figure \ref{fig:uniaa-bench}(a), including 2,229 images in the domain and 3,125 images from the wild.
The aesthetic issues examined in the WILD do not exist in the training of \oursmodel, as indicated by the item highlighted in red in Figure \ref{fig:uniaa-bench}(a).
For each image, there is one question related to perception, with one correct option and 1-3 false options. 
Drawing on recent studies on aesthetic perception~\cite{Yang_2022_CVPR} and the human perception process related to image aesthetics~\cite{li2023theme}, the UNIIAA-QA subset addresses six dimensions through its questions: \textbf{Content and Theme, Composition, Color, Light, Focus and Sentiment}.

\textbf{Content and Theme} are the first aspects that need to be understood when analyzing the aesthetics \cite{leder2004model,li2023theme}. 
Question design includes asking about photo theme, subject type, content, and influence on aesthetics.
\textbf{Composition} is a critical indicator in characterizing the aesthetic qualities of an image \cite{celona2022composition}, which structures the visual story and enhances the aesthetic appeal.
Within visual arts, the visual elements of an image are interdependent through composition~\cite{liu2019modeling}, relying on each other to collectively showcase the aesthetic qualities of the entire image.
Composition contains multiple categories, such as the rule of thirds, diagonal, symmetry, and centering.
\textbf{Color} is considered more perceptually distinct and relevant than other aesthetic features~\cite{he2023thinking} since the human eye can directly perceive different wavelengths of light and convert them into distinct color signals.
\oursbench examines color scheme, color combination, and color harmony.
\textbf{Light} affects aspects of the image such as brightness, contrast, and saturation, which can create a sense of visual hierarchy and depth.
The perception of light includes the lighting quality, the type of luminance source, and the lighting tone.
\textbf{Focus} highlights the important part of the image. A good focus can help the viewer better understand the story of the image and the emotions conveyed, thus improving the aesthetic quality of the image.
\textbf{Sentiment} plays a crucial role in  visual aesthetic~\cite{feng2023probing,leder2004model}. 
It encompasses emotional reactions, attractiveness, and understanding of atmosphere in aesthetic experiences.

The \oursframe-QA utilizes three question types commonly used in human queries when interacting with MLLMs \cite{chen2023egoplan, wu2023towards}: \textbf{"Yes-or-No" questions} requires a positive or negative response (e.g., Does this photo employ the rule of thirds in its composition?). \textbf{"What" questions} examine more comprehensive aesthetic perception, such as "What type of composition is used in this photo?". \textbf{"How" questions} are used to evaluate more details about the grade and production of aesthetic feeling in the image, such as "How does the use of symmetry composition affect photo aesthetic quality?".
\begin{table*}[!t]
\caption{Evaluation results on the \textbf{Aesthetic Perception} ability. We highlight the best result of each subclass with bold and underline the second, third, and fourth best sub-optimal results. CS means the model is the close source. \textit{*GPT-4V may refuse to answer some questions, and the accuracy is calculated based on the questions it answers.}}  \label{AesP}
\renewcommand{\arraystretch}{1.65}
\fontsize{8.5pt}{7pt}\selectfont %
\centering
\setlength{\tabcolsep}{0.9mm}{
\begin{tabular}{l|cccccc|ccc|cc|cc}
\toprule
\multirow{2}*{MLLM}    & \multicolumn{6}{c|}{\textbf{Aesthetic Attributes}} 
& \multicolumn{3}{c|}{\textbf{Question Types}} & \multicolumn{2}{c|}{\textbf{Image Sources}}   & {\multirow{2}*{\textbf{Overall}}}& {\multirow{2}*{\textbf{Rank}}}\\
\cmidrule(lr){2-7}\cmidrule(lr){8-10}\cmidrule(lr){11-12}
&\emph{Con. The.} &\emph{Comp.} &\emph{Color} &\emph{Light} &\emph{Focus} &\emph{Sent.} &\emph{Yes-No} &\emph{What} &\emph{How} &\emph{In-Domain } &\emph{Wild}    \\
\midrule 
\rowcolor{gray!25}
\emph{Random guess} &36.15\% &33.67\% &39.43\% &39.62\% &38.07\% &41.72\%  &50.00\% &32.93\% &28.68\% &33.42\% &41.05\% &37.80\% &$\square$  \\
\midrule 
\midrule 
\rowcolor{lightgray!25}
\emph{Junior-Level Human} &86.66\% &82.24\% &80.48\% &68.14\% &79.39\% &72.41\%  &79.91\% &81.95\% &72.78\% &75.41\% &81.57\% &79.01\% &$\triangle$  \\
\rowcolor{lightgray!25}
\emph{Seinor-Level Human} &91.23\% &88.44\% &80.68\% &89.34\% &87.18\% &90.30\%  &89.54\% &90.08\% &79.91\% &79.81\% &92.80\% &87.39\% &$\triangle$  \\
\midrule 
\oursmodel           &\underline{84.88\%} &\underline{66.82\%} &\textbf{81.58\%} &\textbf{75.68\%} &\textbf{81.98\%} &\textbf{82.75\%} &\textbf{81.64\%} &\textbf{75.83\%} &\textbf{78.70\%} &\textbf{78.78\%} &\textbf{78.27\%} &\textbf{78.48\%} &\textbf{1}\\
\midrule 
$\text{GPT-4V}^{\boldsymbol{*}}$(CS)          &\textbf{85.06\%} &\textbf{70.14\%} &\underline{73.11\%} &\underline{72.10\%} &\underline{73.19\%} &\underline{81.54\%} &\underline{78.99\%} &\underline{74.82\%} &\underline{67.01\%} &\underline{70.56\%} &\underline{77.33\%} &$\underline{74.74\%}$
&\textbf{2}\\
mPLUG-Owl2          &\underline{80.30\%} &56.48\% &\underline{66.27\%} &\underline{64.50\%} &\underline{67.90\%} &\underline{74.57\%} &\underline{71.34\%} &\underline{67.58\%} &\underline{61.11\%}  &\underline{61.19\%} &\underline{71.55\%} &\underline{67.24\%} &\textbf{3}\\
Gemini Pro Vision (CS)   &79.80\% &\underline{56.70\%} &\underline{64.96\%} &61.38\% &\underline{69.10\%} &\underline{76.94\%} &{66.60\%} &64.24\% &\underline{66.19\%} &\underline{69.89\%} &66.60\% &\underline{66.61\%} &{4}\\
ShareGPT-4V          &77.38\% &\underline{57.61\%} &59.76\% &\underline{64.37\%} &60.82\% &69.18\% &\underline{68.90\%} &\underline{68.29\%} &48.67\% &52.89\% &\underline{71.33\%} &63.65\% &{5}\\
Q-Instruct          &\underline{80.30\%} &48.03\% &59.06\% &59.82\% &62.08\% &70.91\% &66.67\% &58.99\% &60.12\% &60.39\% &62.88\% &61.84\% &{6}\\
InternLM-XComposer-VL 
&71.79\% &54.04\% &59.16\% &53.71\% &65.36\% &67.89\% &67.21\% &63.95\% &44.66\% &51.82\% &65.41\% &59.75\% &{7}\\
LLaVA-1.5           &75.10\% &48.87\% &55.47\% &59.43\% &62.02\% &70.26\% &65.11\% &61.63\% &45.64\% &50.52\% &64.80\% &58.85\% &{8}\\
InstructBLIP      &66.11\% &46.13\% &54.76\% &39.69\% &50.36\% &58.39\% &55.70\% &53.09\% &44.25\% &49.32\% &53.59\% &51.81\% &{9}\\
MiniGPT-v2       &56.42\% &37.50\% &45.45\% &44.34\% &50.75\% &62.07\% &54.31\% &49.11\% &37.68\% &43.20\% &51.46\% &48.02\% &{10}\\
OtterHD       &51.97\% &41.73\% &40.54\% &42.26\% &52.72\% &55.60\% &56.62\% &44.36\% &36.24\% &41.18\% &50.43\% &46.58\% &{11}\\
LLaVA               &38.25\% &43.98\% &45.95\% &39.92\% &32.41\% &53.45\% &52.75\% &37.52\% &30.93\% &40.15\% &41.60\% &40.99\% &{12}\\
Qwen-VL         &37.61\% &28.53\% &38.44\% &35.24\% &33.33\% &29.96\% &43.20\% &32.05\% &26.08\% &34.37\% &33.97\% &34.14\% &{13}\\
\bottomrule
\end{tabular}}
\end{table*}

\subsection{Aesthetic Description}
We construct the \textbf{\oursframe-Describe} subset to examine the aesthetic language description ability of MLLMs.
\oursframe-Describe is a golden aesthetic description dataset, consisting of a single long (averaging 100 words) expert-provided description for each image.
The dataset consists of 501 images, categorized into low, medium, and high aesthetic quality with 151, 150, and 200 images respectively.
The description includes a concise and detailed analysis of the aesthetic attributes and photographic techniques used in the image, such as composition, color, focus, lighting, and sentiment. 
Improvement suggestions and overall aesthetic conclusions are summarized at the end of the description.
All descriptions have been written by aesthetic experts and manually reviewed.
Referring to these golden text descriptions, we can measure the quality of MLLMs output using a single-modal GPT \cite{wu2023qbench}, evaluated along three dimensions: \textbf{Completeness}, \textbf{Preciseness}, and \textbf{Relevance}.
The prompt words used to generate descriptions for MLLM are as follows:

\textbf{\textit{User: Describe and evaluate the aesthetic attribute in detail.}}

For images with medium and low aesthetic quality, we prompt MLLM to generate additional aesthetic improvement suggestions:
\textbf{\textit{Give suggestions for improvement.}}
The specific GPT evaluation prompts can be found in Appendix \ref{edad}.

\subsection{Aesthetic Assessment}\label{sec:aa}
The third task tests the zero-shot capability of MLLMs to provide quantitative scores of the overall aesthetic of images.
Recent explorations \cite{wu2023qbench} have observed that LMMs exhibit similar behavior patterns to humans when instructed to score, preferring to respond defined levels, such as \textit{good} and \textit{poor}. 
Using the prompt,

\textbf{\textit{User: Rate this image from an aesthetic perspective.}}

\textbf{\textit{MLLM: The aesthetic quality is [Score Token].}}

MLLMs can provide log probabilities (logits) of the score token.
The standard rating level texts~\cite{series2012methodology}, are \textit{bad, poor, fair, good, excellent}.
Inspired by Q-bench\cite{wu2023qbench} and Q-align\cite{wu2023q}, we employ softmax pooling on the logits of rating level words of the score token, to obtain their probabilities. Figure \ref{fig:uniaa-bench}(c) gives an assessment example.

The quantified scores $S_i$ for the five rating-level words:
\begin{center}
 \textbf{\textit{Excellent: 5, Good: 4, Fair: 3, Poor: 2, Bad: 1}}   
\end{center}

The probability calculation for each rating-level word is as follows:
\begin{equation}
    p_{r_i} = \frac{e^{\mathcal{L}_{r_i}}}{\sum_{j=1}^{5} {e^{\mathcal{L}_{r_j}}}},
\end{equation}
where $r_i$ is the rating-level word, and $\mathcal{L}_{r_i}$ represents the logits of the rating word output by MLLMs.
We obtain an aesthetic quantification score of the image using MLLMs by multiplying the corresponding quantified score of the five rating-level adjectives by their respective probabilities:
\begin{equation}
    \mathrm{S_{MLMM}}=\sum_{i=1}^5 p_{r_i} S_i
    \label{eq:5}
\end{equation}
$\mathrm{S_{MLMM}}$ and the ground-truth scores will be used to calculate Spearman’s rank correlation coefficient (SRCC) and Pearson linear correlation coefficient (PLCC).
The higher these two metrics are, the better the scoring ability of MLLMs.
We select four mainstream IAA dataset sources: AVA, PARA, TAD66K, and Flick-Aes~\cite{ren2017personalized}, of which the training set of the first two are used in the training process, while the latter two are not.
They are depicted in Appendix \ref{detail_assessment_datasets}.
It is worth noting that although we use the aesthetic information annotations of AVA and PARA during the training process, we do not use their MOS for training.
So it is a zero-shot assessment task.

\section{Experiments}
We select LLaVA-1.5-7B~\cite{liu2023improved} as the base model, which contains a ViT-L image encoder, a LLaMA-2-7B, and an MLP connector. 
Detailed training information is in the Appendix \ref{training_detail}.
In \oursbench, we assess 13 variants on the 10 most current and competitive open-source MLLMs, as well as GPT-4V and Gemini-Pro-Vision.
We also report the perception results of human aesthetic judgment.

\subsection{Results on Aesthetic Perception}
We assess the aesthetic perception of MLLMs by testing their capability to provide accurate responses to uncomplicated natural language queries related to aesthetic attributes.
Table \ref{AesP} summarizes the outcomes.
It is pleasing to note that the majority of MLLMs can significantly surpass random guessing on all aesthetic attributes. 
Except for Qwen-VL, it will answer "uncertain" for some of the questions about aesthetic choices. 
We provide its response examples in the Appendix \ref{epr}.
Given that these general-purpose models did not receive much explicit training on visual aesthetic attributes, these findings demonstrate their strong potential for aesthetics.
By comparing \oursmodel with the original LLaVA-1.5, it can be found that using aesthetic fine-tuning data obtained by the IAA conversion pipeline can significantly improve the aesthetic perception ability of MLLM.
There is an increase of over 10\% in almost all aesthetic dimensions. 
It is worth noting that the accuracy of composition has been improved from less than 50\% to 66.82\%.
In terms of question types, the largest improvement is in the "How" type of questions, which increases by 30\%.
The simultaneous improvement of \oursmodel in both in-the-wild and in-domain data proves its generalization of aesthetic perception ability.
This finding demonstrates that aesthetic abilities are transferable and can be well generalized to unseen images.
Compared with GPT-4V, we achieve a leading performance in four out of six dimensions except for the composition and content and have an overall accuracy higher than GPT-4V (78.48\% vs. 74.74\%).
In addition to our model, two closed-source models have a high visual aesthetic perception ability in mainstream MLLMs. 
There is no doubt that GPT-4V is the strongest, with Gemini Pro ranking fourth among all models and only slightly behind mPLUG-Owl2.
The best in open-source MLLM is mPLUG-Owl2, with its perceptual accuracy surpassing ShareGPT-4V by 3.59\%.
The most challenging perceptual dimension for MLLM is \textbf{Composition}, with most models not exceeding 60\%. 
This may be due to the difficulty of the composition appreciation task and the diversity of composition techniques.
Compared to "what" and "yes/no" questions, MLLMs generally have low accuracy in answering "how" questions, indicating that their ability to rate aesthetics needs to be improved.

\textbf{MLLMs VS. Human.} Although leading other models, \oursmodel still lags behind Junior-level Humans slightly (78.48\% VS. 79.01\%).
Despite its prowess, there is still a way to go for
\oursmodel before it can match the overall proficiency of the Senior-level Human (with experiences on visual aesthetic tasks, 10\% better than \oursmodel).
The aspect in which the largest gap between MLLMs and humans exists is composition.
The comparative results with human evaluation indicate that MLLMs still have a long way to go in terms of visual aesthetics, and it is necessary to continue enhancing the visual aesthetic ability of MLLMs.

\subsection{Result on Aesthetic Description}
In terms of descriptive ability, as shown in Table \ref{AesD}, GPT-4V achieves the highest performance and \oursmodel takes the lead in completeness and preciseness.
As accurate description is the most difficult, all MLLMs have not achieved a score of 1 or higher in preciseness.
Instead, They achieve good scores in completeness, with scores higher than 0.8.
Overall, only the GPT-4V, \oursmodel, and ShareGPT-4V score over 3 out of 6. 
All the MLLMs fail the \oursframe-Describe test (3.6/6.0).
MLLMs have relatively limited and basic abilities in providing aesthetic descriptions.
Detailed examples and qualitative analysis can be found in Appendix \ref{app:qad}.
\begin{table}[!t]
\caption{Evaluation on the Aesthetic Description. Due to the high cost and difficulty, human results are not be provided.} \label{AesD}
\renewcommand{\arraystretch}{1.6}
\fontsize{8.5pt}{7pt}\selectfont %
\centering
\setlength{\tabcolsep}{1.4mm}{
\begin{tabular}{l|ccc|c}
\toprule
 MLLM  &\emph{Comp.} &\emph{Prec.} &\emph{Rele.} & \textbf{Overall}\\
\midrule 
\oursmodel          &\textbf{1.581} &\textbf{0.821} & \underline{1.046}     &\underline{3.448} \\
\midrule 
GPT-4V (CS)             &\underline{
1.448} &\underline{0.754} &\textbf{1.372} &\textbf{3.574} \\
ShareGPT-4V         &\underline{1.427} &\underline{0.722} &\underline{0.978}     &\underline{3.127} \\
LLaVA-1.5           &1.271 &\underline{0.709}    &0.948 &2.928 \\
LLaVA               &1.287 &0.673     &0.956 &2.916 \\
mPLUG-Owl2          &\underline{1.305} &0.671    &0.932 &2.908 \\
Gemini Pro Vision (CS)  &1.269 &0.623 &\underline{0.972}     &2.864 \\
OtterHD               &1.112 &0.529     &0.894 & 2.535 \\
InternLM-XComposer-VL &0.996 &0.419 &0.914     &2.329 \\
Q-Instruct          &1.022 &0.327     &0.930 &2.279 \\
MiniGPT-v2          &0.889 &0.353    &0.846 &2.088 \\
InstructBLIP        &0.898 &0.337     &0.796 &2.031 \\

\bottomrule
\end{tabular}}
\end{table}

\begin{table}[!t]
\caption{Evaluation results on the \textbf{Aesthetic Assessment} ability. Metrics are PLCC/SRCC. Seen means using the data in training \oursmodel, while unseen means not using. Due to the enormous quantity, we do not use human-assessment.} \label{AesA}
\renewcommand{\arraystretch}{1.6}
\fontsize{8.5pt}{7pt}\selectfont %
\centering
\resizebox{1\linewidth}{!}{\setlength{\tabcolsep}{1mm}{
\begin{tabular}{l|cc|cc}
\toprule
\multirow{2}{*}{Model}  &\emph{AVA}  &\emph{PARA} &\textbf{TAD66K} &\textbf{Flickr-Aes}\\

  &Seen  &Seen &Unseen &Unseen\\
\midrule 
\oursmodel           &\textbf{0.704/0.713}  &\textbf{0.895/0.864} &\textbf{0.425/0.411} &\textbf{0.751/0.724}     \\
\midrule 
Aes-CLIP            &0.637/0.647 &- &- &-     \\
VILA    &0.664/0.658 &0.635/0.657  &0.372/0.350  & 0.602/0.577\\
\midrule 
Q-Instruct          &0.330/0.313 &0.785/0.746 &0.160/0.137 &0.585/0.560     \\
ShareGPT-4V          &0.321/0.327 &0.571/0.625 &0.186/0.194 &0.554/0.565     \\
mPLUG-Owl2          &0.328/0.353 &0.614/0.628 &0.198/0.215  &0.484/0.509     \\
LLaVA-1.5           &0.281/0.286 &0.477/0.558 &0.158/0.156 &0.456/0.496     \\
LLaVA               &0.257/0.267 &0.422/0.521 &0.145/0.153 &0.408/0.456     \\
\bottomrule
\end{tabular}}}
\end{table}

\subsection{Result on Aesthetic Assessment}\label{raa}
As depicted in \ref{sec:aa}, we use the score token to predict aesthetic scores, and compute the PLCC/SRCC concerning ground truth.
The results are listed in Table \ref{AesA}.
\oursmodel leads zero-shot score performance on all four mainstream IAA datasets, even though it has not been trained on TAD66K and Flick-AES.
It's worth noting that we do not use MOS values during the training.
The result indicates that we can fine-tune MLLMs effectively on Aesthetic Assessment by utilizing aesthetic knowledge transformed from IDCP.
Additionally, we obtain the state-of-the-art (\textbf{SOTA}) performance on AVA and TAD66K by conducting supervised training using the vision encoder of \oursmodel, as shown in Appendix \ref{appendix.ss}.
We achieve a PLCC of \textbf{0.838} on AVA and \textbf{0.553} on TAD66K with our vision encoder and some additional linear layers.

\begin{table}[!t]
\caption{Ablation study. AA represents using aesthetic attribute data obtained from IDCP to optimize MLLM, while AC represents using aesthetic comment/description data. RAW represents using raw comment data without going through IDCP.} \label{ablation}
\renewcommand{\arraystretch}{1.6}
\fontsize{8.5pt}{7pt}\selectfont %
\centering
\setlength{\tabcolsep}{0.8mm}{
\begin{tabular}{cc|ccc|c|cc}
\toprule
  \multirow{2}*{AA}  & \multirow{2}*{AC}  &\multicolumn{3}{c|}{\textbf{\emph{Perception} }} & \multirow{2}*{\textbf{\emph{Desc.}}} & \multirow{2}*{TAD66K}& \multirow{2}*{Flickr-Aes}\\
  & &In-domain & Wild& Overall & & & \\
\midrule 
$\checkmark$ &$\checkmark$ &78.78\% &78.27\% &78.48\% & 3.448&\textbf{0.43/0.41} &\textbf{0.75/0.72}\\
$\checkmark$ &$\times$ &75.68\% &72.58\%  & 73.87\%&2.511 &0.32/0.30& 0.68/0.66\\
$\times$ &$\checkmark$ &68.13\% &71.60\%  &70.16\% &3.519 &0.39/0.40 &0.63/0.62\\
$\checkmark$ & \emph{RAW} &77.25\% &72.83\% &74.68\% & 2.124&0.32/0.35&0.70/0.72\\ \hline
\multicolumn{2}{c|}{\footnotesize \oursframe-OWL2} & 77.70\% & 77.98\% &77.89\% &3.519 &0.42/0.41 &0.74/0.72 \\ 
\multicolumn{2}{c|}{Freeze VE} & 75.41\% & 77.82\% &76.82\% &\textbf{3.540} &0.42/0.41 &0.73/0.71 \\

\bottomrule
\end{tabular}}
\end{table}

\subsection{Ablation Study}
\subsubsection{IDCP ablation experiments}
To validate the effectiveness of IDCP, we conduct ablation experiments on aesthetic data types, cleaning steps, and additional MLLM, as shown in Table \ref{ablation}.
When using only aesthetic attributes or aesthetic descriptions, a decrease in perception occurs in both cases, with a greater decrease observed in the latter.
This reflects the importance of attributes and comments in improving aesthetic capabilities.
The use of raw comments training led to a 3.8\% decrease in WILD, demonstrating that lacking IDCP cleaning can affect generalization.
Meanwhile, the description ability decreases significantly, indicating the necessity of rewriting the raw data using IDCP.
When any of the IDCP stages is missing, the performance of the model will decrease to a different extent.

 \begin{table}[t]
    \centering
    \caption{Results on Q-bench LLVisionQA dev set.
     QI represents Q-Instruct data for image quality assessment finetuning.}
    \renewcommand{\arraystretch}{1}
    \begin{tabular}{lcccc}
    \toprule
        Method & Acc. \\
    \midrule
        InstructBLIP                &56.72\% \\
        LLaVa-v1.5-7B               &58.66\%  \\
        LLaMA-Adapter-V2            &59.46\% \\
        LLaVA-v1.5-13B              &62.14\% \\
        mPLUG-Owl2                   &62.90\% \\
        InternLM-XComposer-VL       &65.35\% \\
        Qwen-VL-Plus (CS) & 66.04\%\\
        Gemini-pro-vison (CS)   &68.16\%\\
        
    \midrule
    QI-LLaVA-v1.5-7B                &67.09\%\\
    QI-LLaVA-v1.5-13B               &67.42\%\\
    \midrule

    \oursmodel-v1.5-7B                   &\textbf{69.43\%}\\
    \bottomrule
    \end{tabular}
    \label{tab:qbench}
\end{table}

\subsubsection{\oursframe on mPLUG-OWL2}
We obtain the \oursframe-mPLUG-OWL2 using the same data of \oursmodel, and find that its accuracy on the Aesthetic Perception is 77.89\%, which is slightly lower than \oursmodel and better than GPT-4V.
This further proves the generality of IDCP. 
\oursframe-mPLUG-OWL2 also achieves significant improvements in both description and assessment compared to the vanilla model.
\subsubsection{Ablation of Fine-tuning the vision encoder.}
We freeze the weights of vision encoder during training while keeping other training settings consistent. 
The perception is 76.82\%, lower than the original \oursmodel (\textbf{78.48\%}), and a similar trend is observed in assessment.
It shows that IAA tasks require a finer level of visual perception. 
Freezing VE slightly improves aesthetic description.
\subsubsection{Generalization on image quality benchmark.}
Q-bench \cite{wu2023qbench} is a low-level visual benchmark on image quality assessment. 
As shown in Table \ref{tab:qbench},  \oursmodel performs surprisingly well. Without being trained on image quality datasets, it not only outperforms many open-source MLLMs, but also surpasses models trained using Q-instruct, which is specifically designed for image quality.
The consistency result between Q-bench and \oursbench demonstrates the effectiveness of IDCP and the reliability of \oursbench.

\section{Conclusion}
In this work, we introduce \textbf{Un}ified Multi-modal \textbf{I}mage \textbf{A}esthetic \textbf{A}ssessment (\textbf{\oursframe}), a complete framework to enhance and examine the visual aesthetic ability of MLLMs, including \textbf{\oursmodel}~ and \textbf{\oursbench}.
We propose the first-of-a-kind multi-modal IAA Datasets Conversion Paradigm (IDCP).
It permits MLLMs to substantially enhance accuracy in answering queries about visual aesthetic perception and also demonstrates the potential to produce more dependable aesthetic descriptions of images. 
Ultimately, this could serve to alleviate the workload of human experts in performing IAA.
\oursbench examines MLLMs based on three crucial capabilities: the ability to perceive aesthetic attributes, the ability to provide an accurate and thorough aesthetic description, and the capacity to offer a score of aesthetic quality.
The evaluation carried out by \oursbench has yielded two significant findings.
The first finding is that \oursmodel exhibits superior aesthetic abilities compared to other MLLM models, which serves as evidence of the effectiveness of IDCP.
The second finding is that, in comparison to human experts, MLLMs still have a significant distance to traverse in order to become a truly dependable general visual aesthetic assistant.
We sincerely hope that \oursframe can inspire future MLLMs to enhance their aesthetic abilities.

\bibliographystyle{ACM-Reference-Format}
\bibliography{main}

\newpage
\appendix
\section{More Information of IDCP}
\subsection{Data Source Details}\label{appendix:drd}
The sample size of IDCP using five IAA datasets sources is shown in Table \ref{tab:drd}.
Overall Comments provide a general evaluation of the photo. 
Photo Style is an annotation of the photographic and aesthetic style, such as macro photography, motion blur, long exposure, etc.
Binary Attributes are binary annotations of aesthetic attributes, Such as whether there is balance in the elements, and whether the rule of thirds is applied.
Attributes Level is an evaluation of the quality level of the aesthetic attributes. 
Single Attribute Comments are comments about a specific dimension of the photo, and Color Scheme is an annotation of the color combinations in the photo.
\begin{table}[h]
    \centering
    \caption{Datasets resource detail. The total number of data sources is 278,181.}
    \renewcommand{\arraystretch}{1.6}
\fontsize{9pt}{8pt}\selectfont %
    \begin{tabular}{lccc}
    \toprule
        Dataset &Annotation &Size  \\
    \midrule
        \multirow{2}*{AVA} & Overall Commments &26562 \\
                        & Photo style &30957 \\
        \hline
        AADB & Binary Attributes & 30976\\ 
        \hline
        PARA  & Attributes Level &141100 \\
        \hline
        \multirow{2}*{PCCD} & Overall Comments &4202 \\
        & Single Attribute Comments & 27033 \\ \hline
        ICAA17K & Color Scheme & 17351\\ \hline

        \multirow{2}*{\textbf{Overall}} & Aesthetic Attributes & 220384\\
         & Aesthetic Comments & 57797\\ 
\bottomrule
    \end{tabular}

    \label{tab:drd}

\end{table}
\subsection{Prompt of Question Generation on Aesthetic Attributes}\label{appendix-pqgaa}
We first assign the role of an aesthetic expert to ChatGPT and ask it to design a question based on a specific aesthetic dimension. 
Then, while maintaining the same meaning, we ask it to expand the question into multiple different sentence structures.
The following is a prompt for designing a question about composition.

\textit{\textbf{User}: Assuming you are an aesthetic expert, please design a question to ask What kind of composition is used for this photo. Rewrite the question into five different sentences with the same meaning.}

\subsection{Prompt of Description Generation}\label{app:pdg}
Given an image, we use LLaVA to generate its content. 
LLaVA is required to output the image content and spatial information.
It is not allowed to provide any aesthetic-related information to prevent any interference with the real aesthetic annotation.

\textit{\textbf{User}: Provide a detailed description of this image, including the spatial and relative position of the elements in the picture.
Do not output any information related to aesthetics, such as atmosphere, emotion, and the beauty of the scenery, etc.}

We input the annotated comments and the captions generated by LLaVA into GPT and request GPT to rewrite the comments.

\textit{\textbf{User}: Assistant is a professional aesthetic critic.}

\textit{ The assistant is a highly professional individual who possesses a remarkable eye for aesthetics and aligns well with the views of the general public. Instructions:
- You're given a caption of an image and some comments on the image from a photography website.}

\textit{ - Step1: Rewrite the comments into professional aesthetic comments and ignore those comments that are not highly related to aesthetics or only expressions of personal affection. 
Please be objective and do not use colloquial words.}

\textit{ - Step2: Summarize the rewritten aesthetic comments and your answers in Step2 into one complete aesthetic comment written by a professional aesthetic critic.
You may refer to the image caption as though you are truly seeing this image, but please focus solely on the aesthetic-related content.}

\textit{ When the image caption conflicts with the given comments, follow the comments.
Do not imagine and give irrelevant or groundless responses regarding the given comments.}

\subsection {Examples of IDCP Data}\label{app:idcp_example}
Following are some examples of training data obtained through IDCP transformation.

\noindent\textbf{Aesthetic Attributes\\}
\noindent\textit{\textbf{AVA}: Question: What kind of photography style is this image in? Answer: Soft Focus.}

\noindent\textit{\textbf{AADB}:  Question: Are the elements in this image balancing?  Answer: No.}

\noindent\textit{\textbf{PARA}:  Question: How would you rate the composition of this image?  Answer: The image has a fair score of composition.}

\noindent\textit{\textbf{ICAA17K}:  Question: What is the color scheme employed in this picture?  Answer: Complementary.}

\noindent\textit{Question: What complementary colors are used in this photo?  Answer: Orange and Violet.}

\noindent\textbf{Aesthetic Comments\\}
\noindent\textit{\textbf{AVA}: Question: What is your overall impression of this image from an aesthetic viewpoint?\\}
\noindent\textit{Answer: The image seems to be the result of well-coordinated, meticulous work that features impressive captures of the lunar eclipse across different stages. The successful inclusion of background stars further enriches the image's overall visual appeal and depth. Expert editing appears to have been utilized proficiently, contributing substantially to the enchanting allure of the artwork. However, a minor flaw can be noted on the eighth moon, indicating room for further refinements. The work invokes a blend of admiration and curiosity, leaving the observer to ponder whether this breathtaking display was captured entirely in nature or whether digital enhancements played a role. Nonetheless, the image stands out as a favorite due to its unique and striking aesthetic elements.}

\noindent\textit{\textbf{PCCD}: Question: Summarize your aesthetic evaluations of this image as if you are a professional aesthetic critic.\\}
\noindent\textit{Answer: The composition of the image is well-executed, with the baby girl placed in the center and occupying a significant portion of the frame. The use of a white blanket as the background creates a serene and soothing atmosphere. The pink bow in the baby's hair adds a touch of color and serves as the main focal point of the image. The lighting, with its rosy effect, enhances the overall aesthetic appeal, giving the image a gentle and warm tone. However, the image lacks a strong emotional impact and fails to capture the viewer's attention.}

\noindent\textit{\textbf{PCCD}: Question: How would you evaluate the aesthetic expression of the \textbf{use of camera} of this image?\\}
\noindent\textit{Answer: The photographer made skillful choices in the use of the camera for this image. The angle chosen provides an interesting perspective, allowing us to fully appreciate the details of the subject. The f-stop and shutter speed selection contribute to the clean and well-captured image, ensuring that the details are sharp and clear.}

\subsection{Hyper-parameters of Training}\label{training_detail}
Table \ref{tab:hyperparam} contains the hyper-parameters utilized for tuning the IAA instructions in LLaVA-v1.5-7B and mPLUG-OWL2 using 8 NVIDIA V100 GPU. The fine-tuning took a total of 16 hours for \oursmodel and 22 hours for \oursframe-mPLUG-OWL2.
We conduct a third stage of fine-tuning solely with converted aesthetic instruct tuning data to obtain the \oursmodel for 2 epochs with a learning rate of 2e-5 and a batch size of 128.
\begin{table}[htbp]
\centering
\caption{
\textbf{Hyper-parameters} of \textit{IAA visual instruction tuning} on \oursmodel and \oursframe-mPLUG-OWL2, the same as original baselines.
}
\resizebox{\linewidth}{!}{
\begin{tabular}{l| c c}
\toprule
Hyper-parameter &  \textbf{\oursmodel} &  \textbf{\oursframe-mPLUG-OWL2} \\
\midrule
ViT init. & {CLIP-L/14-336 } & CLIP-L/14-448  \\
LLM init. & \multicolumn{2}{c}{LLaMA2-7B} \\
image resolution         & $336\times 336$ & $448\times 448$ \\
batch size & \multicolumn{2}{c}{128} \\
lr max & \multicolumn{2}{c}{2e-5} \\
lr schedule & \multicolumn{2}{c}{cosine decay} \\
warmup epochs & \multicolumn{2}{c}{0.03} \\
weight decay & \multicolumn{2}{c}{0} \\
gradient acc.            & \multicolumn{2}{c}{1} \\
numerical precision      & \multicolumn{2}{c}{$\mathtt{Float16}$} \\
epoch & \multicolumn{2}{c}{2} \\
optimizer & \multicolumn{2}{c}{AdamW} \\
optimizer sharding       & \multicolumn{2}{c}{\checkmark} \\
activation checkpointing & \multicolumn{2}{c}{\checkmark} \\
deepspeed stage & \multicolumn{2}{c}{3} \\
\bottomrule
\end{tabular}
}

\label{tab:hyperparam}
\end{table}

\section{Details on \oursbench Evaluation}
\subsection {Examples of \oursframe-QA}
Here we present examples of \oursframe-QA questions with different aesthetic attributes.

\noindent\textbf{Content and Theme\\}
\noindent\textit{Question: What is the object of interest in this photo?}
\noindent\textit{Candidates: A. Landscape B. Portrait C. Still life D. Animal}

\noindent\textbf{Composition\\}
\noindent\textit{Question: What is the specific composition method used in this image?}
\noindent\textit{Candidates: A. Diagonal composition B. Center composition C. Triangle D. Symmetrical composition}

\noindent\textbf{Color\\}
\noindent\textit{Question: What is the color saturation of the picture?}
\noindent\textit{Candidates: A. Oversaturated B. Appropriate C. Undersaturated}

\noindent\textbf{Light\\}
\noindent\textit{Question: What is the source of light in this image?}
\noindent\textit{Candidates: A. Natural light B. Artificial light}

\noindent\textbf{Focus\\}
\noindent\textit{Question: What type of shot is this photo?}
\noindent\textit{Candidates: A. Long shot B. Medium shot C. Close-up}

\noindent\textbf{Sentiment\\}
\noindent\textit{Question:  Based on the color and content of this picture, what emotion does this photo evoke from you? }
\noindent\textit{Candidates:
A. The character's demeanor and the surrounding atmosphere in this image evoke a sense of longing and nostalgia.
B. This picture resonates with a sense of introspection and contemplation due to the character's actions and the environment, devoid of any traces of regret or sadness.
C. The character's actions and environment bring about a feeling of melancholy and regret.
D. Despite the gloomy mood implied by the character's actions and environment in the picture, a feeling of peacefulness and serenity suffuses throughout.}

\subsection {Example of \oursframe-Describe}
\noindent\textit{Question: Describe and evaluate the aesthetic attributes in detail. Give suggestions for improvement.\\}
\noindent\textit{Answer: The upward angle captures the grand structure and spatiality of the urban architecture. The composition is relatively neat and harmonious, with the top of the two buildings forming a diagonal composition, giving the image a certain sense of depth. However, there is a large empty space in the middle, making the image slightly unbalanced. The overall color is dark and low in saturation, lacking color coordination and contrast, making the image dull and uninteresting. The overall brightness is insufficient, lacking layers and details. To improve, the shooting position needs to be changed to reduce the empty sky in the middle of the frame, while emphasizing the diagonal composition to make the image more vivid and layered. The saturation of colors needs to be increased to make the image more vibrant and lively. Lastly, overall exposure needs to be increased to enhance the layers and image details. This is a below-average architectural photography photo.}

\subsection {Example of \oursframe Assessment Dataset}\label{detail_assessment_datasets}
\begin{figure}
    \centering
    \includegraphics[width=0.48\textwidth]{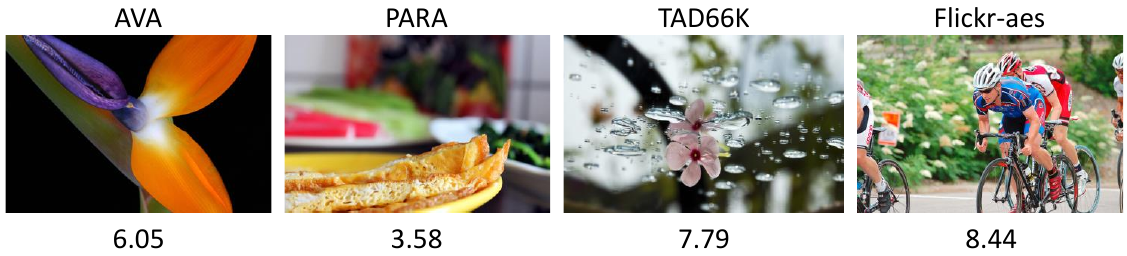}
    \caption{Image examples of four datasets in Aesthetic Assessment. Below the images are the corresponding aesthetic scores.}
    \label{fig:aa_dataset}
\end{figure}

There are 19,928 samples in the AVA test set.
The number of samples in the test set of PARA, TAD66K, and Flickr-aes is 14,079, 3,000, and 4,737, respectively.
The aesthetic assessment datasets image samples are shown in Figure \ref{fig:aa_dataset}. The maximum aesthetic score is 10 for AVA, TAD66K, and Flicker-aes, while it is 5 for PARA.

\subsection{Evaluation Details for Aesthetic Perception}
For the majority of the models, we use multiple-choice prompts as follows,

\noindent\textit{{ How would you describe the color saturation of this picture? {\tt(Question)} [IMAGE\_TOKEN] {\tt(Image)} \\ Choose between one of the options as follows:\\  A. Oversaturated {\tt{(Correct)}}  B. Undersaturated{\tt(Wrong)}  C. Appropriate{\tt(Wrong)}}\\\textcolor{red}{\#Answer:}}

\subsection{Exceptions Perception Responses.}\label{epr}

For the InstructBLIP, although it has aesthetic perception capabilities, it usually provides a paragraph in response instead of options. 
To deal with this, we use chatGPT to determine its choice based on its response,

\textit{  As a language expert, please complete the following task. You are now an answer selection expert, and I will provide you with a question with several options, as well as a target sentence. Please return the alphabet of the option with the highest probability of matching this target sentence. Given questions with options and the target sequence [MLLM ANSWER]. Please output your responses in the form of a dictionary {"maximum probability": "xxx"}, where xxx is A or B or C or ...}

For Qwen-VL, it may present options that do not exist or the response 'information is insufficient', and we consider this to be a wrong answer,

\noindent\textit{{ How would you describe the color saturation of this picture? {\tt(Question)} [IMAGE\_TOKEN] {\tt(Image)} \\ Choose between one of the options as follows:\\  A. Oversaturated {\tt{(Correct)}}  B. Undersaturated{\tt(Wrong)}  C. Appropriate{\tt(Wrong)}}\\\textcolor{red}{\#Answer: D. Not enough information}} 

Regarding GPT-4V, it occasionally encounters unknown errors or refuses to respond. We have made repeated attempts and still encountered 500+ such incidents. Therefore, we calculated its accuracy in answering questions.

\subsection{Evaluation Details for Aesthetic Description}\label{edad}
\textbf{Settings for GPT Evaluation:} 

Due to the inherent variability of GPT, responses to identical prompts may not always be definitive. In order to alleviate the impact of these situations on our evaluation, we have adopted an \textbf{average pooling} strategy over 5 rounds. This involves presenting the same prompt as defined in the following templates five times, and averaging the results of GPT's answers to obtain the final outcome. By using this method, we can effectively alleviate the unpredictability of GPT, resulting in a more accurate score.

\noindent\textbf{Prompt Templates for GPT Evaluation:}

\noindent\textit{  \#System:~As a language expert, please complete the following task.} 

\noindent{\textbf{Completeness.} \textit{\small\#User: Evaluate whether the description [MLLM\_DESC] contains the dimensions of aesthetic, including composition, color, lighting, focus and suggestions, etc., in the reference description [GOLDEN\_DESC]. \\Please rate score 2 for completely or almost completely including aesthetic dimensions, 0 for not including at all, and 1 for including part of the dimensions or similar description.. \\Please only provide the result in the following format: Score:}} \\
{ \textbf{Preciseness.} \textit{\small\# The precision metric punishes highly controversial aesthetic perspectives that output contrasts with the reference, e.g., positive for negative evaluations, good composition for messy composition, harmonious colors for abrupt colors, appropriate lighting for inappropriate lighting, high quality for low quality, colorful for monotonous.\\
Evaluate whether [MLLM\_DESC] precisely reflects reference [GOLDEN\_DESC].  \\
Please rate score 2 for a totally no controversial aesthetic description, 1 for less controversial aesthetic description than the matched description, and 0 for more controversial aesthetic description than the matched.\\
Please only provide the result in the following format: Score: }} \\                  
{ \textbf{Relevance.} \textit{\small\#User: Evaluate whether the [MLLM\_DESC] is relevant to the aesthetic evaluation, aesthetic attributes and aesthetic terminology. Aesthetic attributes include composition, color, lighting, focus, sentiments, and suggestions for improvement.   \\Please rate score 2 for completely relevant, with no content unrelated to aesthetics; 1 for partly relevant, with a small amount of content unrelated to aesthetics; 0 for a large amount of content unrelated to aesthetics, even irrelevant. \\Please only provide the result in the following format: Score:}} \\ 

\begin{figure}
    \centering
    \includegraphics[width=0.2\textwidth]{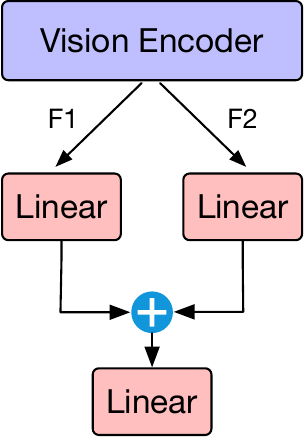}
    \caption{Supervised training model of MOS prediction.  F1 represents the features of the last layer of the Vision Encoder, while F2 represents the features of the second-to-last layer.}
    \label{fig:mos_prediction}
\end{figure}
\begin{table}[t]
    \centering
        \caption{Results on AVA and Tad66K datasets. \best{Blue} and \second{black} numbers in bold represent the best and second best respectively. Metrics are PLCC/SRCC.}
    \renewcommand{\arraystretch}{1.6}
\fontsize{9pt}{8pt}\selectfont %
    \begin{tabular}{lcccc}
    \toprule
        Method &AVA &TAD66K \\
    \midrule
        NIMA \cite{talebi2018nima}       &0.636/0.612 &0.405/0.390 \\
        ALamp\cite{ma2017lamp}                            &0.671/0.666 &0.422/0.411 \\
        HGCN\cite{she2021hierarchical}                             &0.687/0.665 &0.493/0.486 \\
        MaxViT \cite{tu2022maxvit}                      &0.745/0.708 &0.518/0.490 \\
        MUSIQ \cite{Ke_2021_ICCV}                       &0.738/0.726 &0.517/0.489 \\
        MLSP \cite{Hosu_2019_CVPR}           &0.757/0.756 &0.508/0.490 \\
        TANet \cite{he_2022_ijcai}                      &0.765/0.758 & 0.531/0.513 \\
        DAT \cite{xia2022vision}     &0.739/0.738 &0.527/0.499 \\
        EAT \cite{he2023eat}  &\second{0.814/0.803}  &\second{0.546/0.517}\\
    \midrule
        Ours                                           &\best{0.838/0.840} &\best{0.553/0.521} \\
    \bottomrule
    \end{tabular}

    \label{tab:mos_prediction}

\end{table}
\section{Supplemental experiment}
\subsection{Result on Supervised Scoring}\label{appendix.ss}
To further enhance the scoring capability on specific datasets, we extract the vision encoder of \oursmodel. 
We use the features from its last two layers, each of which is separately projected through a linear layer, then added together and passed through a score prediction head.
The entire training model is shown in Figure \ref{fig:mos_prediction}.
We test its performance on AVA and Tad66K, as shown in Table \ref{tab:mos_prediction}.
We use a learning rate of 4e-6, a batch size of 32, a middle dimension of 4096 for the linear layers, and 40 training epochs.
It is evident that even compared with the state-of-the-art EAT\cite{he2023eat}, our model still achieves superiority, further proving the capability of \oursmodel on IAA.

\subsection{Qualitative Analyse on Perception}\label{app:qap}
\begin{figure}
    \centering
    \includegraphics[width=0.48\textwidth]{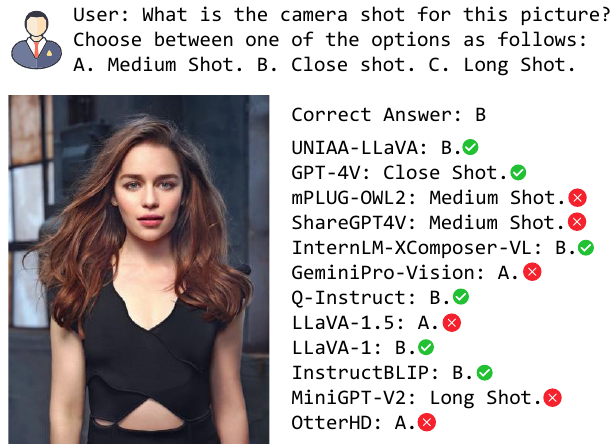}
    \caption{Answer examples of MLLMs on 'Focus' perception.}
    \label{fig:qad}
\end{figure}

In Figure \ref{fig:qad} , we show qualitative examples of MLLM responses on questions in the \oursframe-QA. Some of the MLLM still have inadequate spatial perception of the camera shot.

\subsection{Qualitative Analyse on Description}\label{app:qad}
\begin{figure*}
\centering
\includegraphics[scale=0.9]{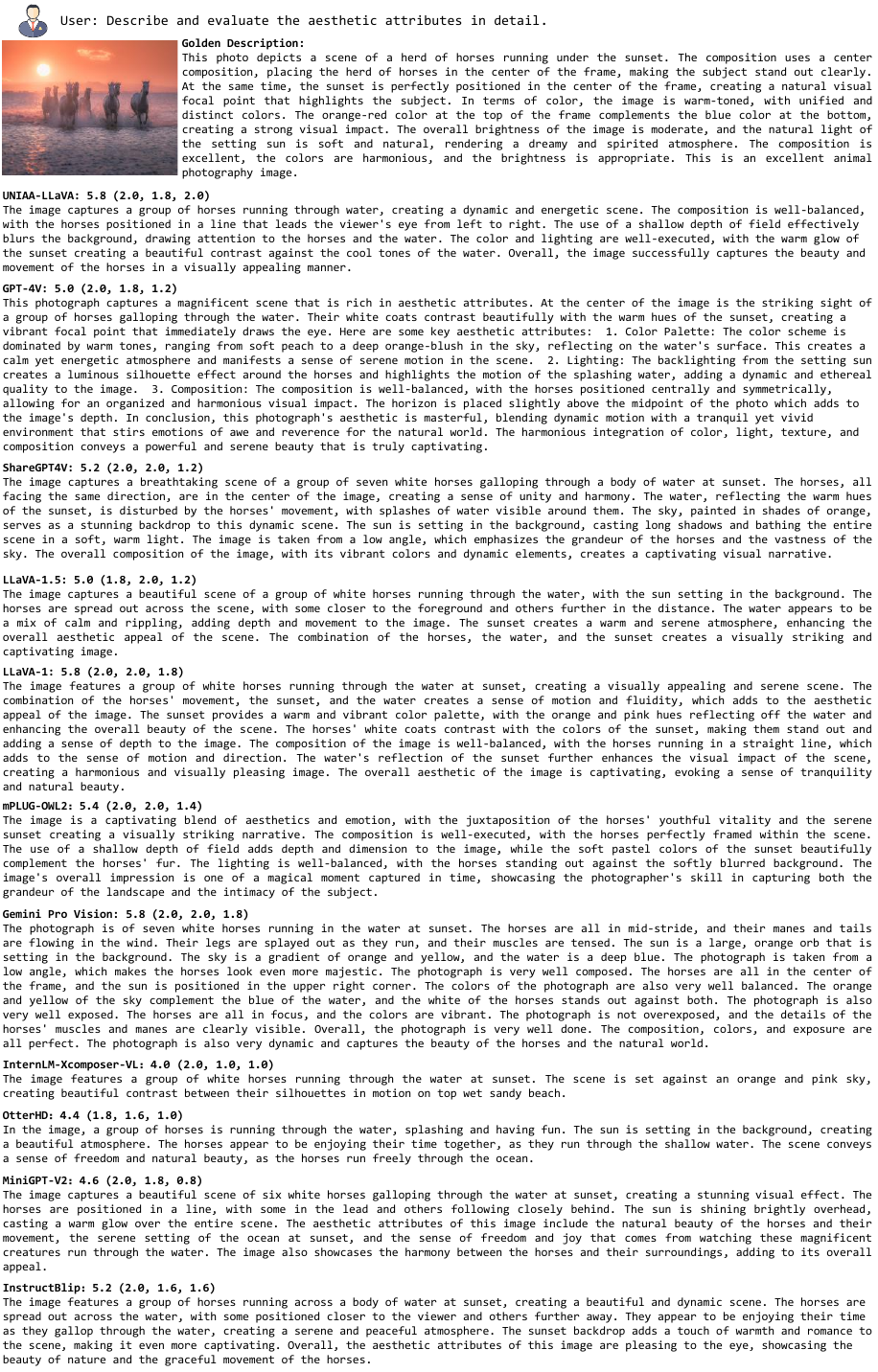}
\caption{Description examples on the high-aesthetic image of MLLMs.}
\label{fig:high_describe_exmple}
\end{figure*}

\begin{figure*}
\centering
\includegraphics[scale=0.88]{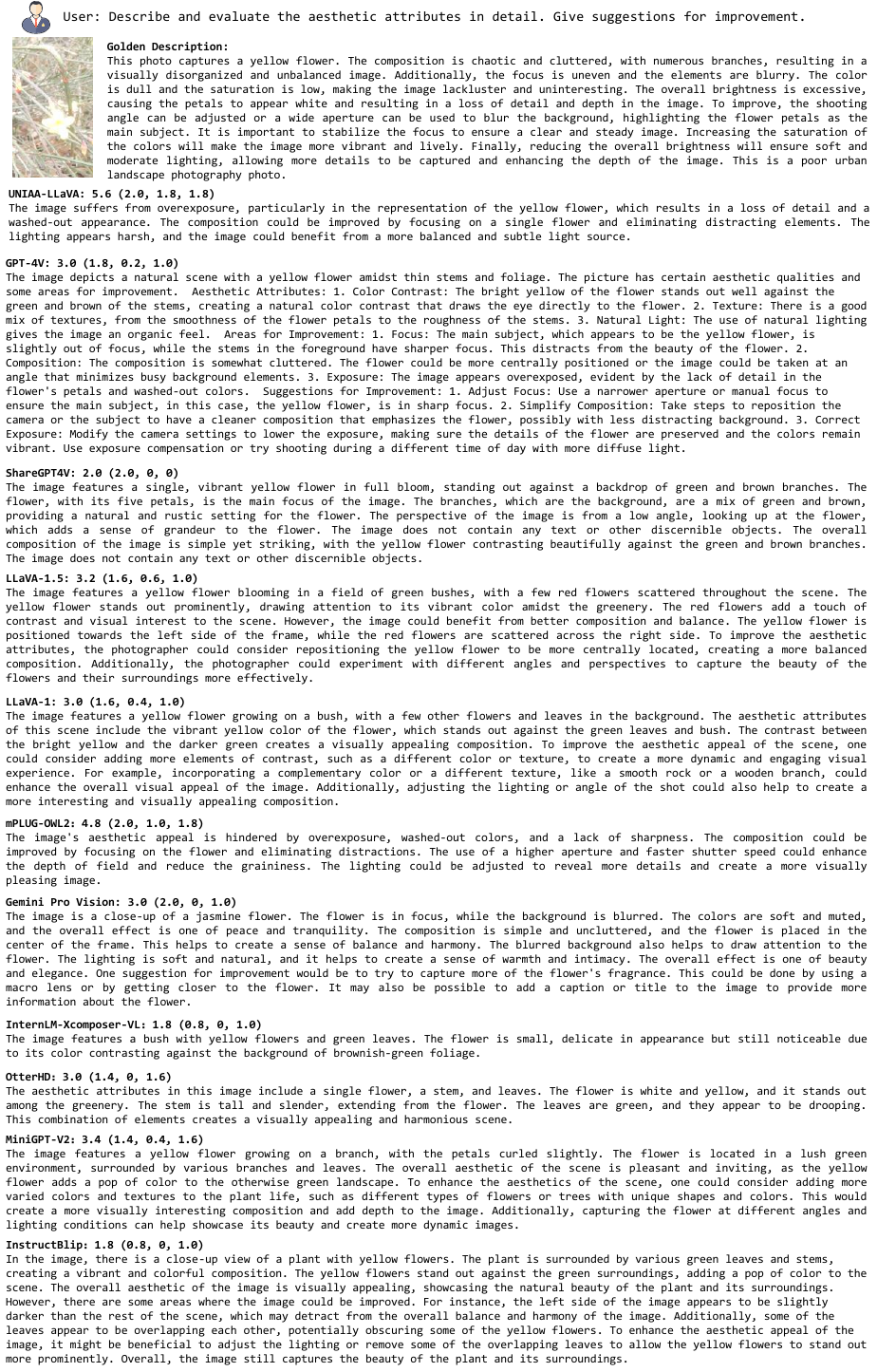}
\caption{Description example on the low-aesthetic image of MLLMs.}
\label{fig:low_describe_exmple}
\end{figure*}

As shown in Figure \ref{fig:high_describe_exmple} and \ref{fig:low_describe_exmple}, we have presented examples of MLLM descriptions of high and low aesthetic quality.
As shown in Figure \ref{fig:high_describe_exmple}, when MLLMs evaluate high aesthetic quality images, their descriptions are  complete, accurate, and relevant, and their scores are high.
\oursmodel uses more aesthetic terms such as well-balanced composition, shallow depth of field, and color contrast, providing a professional description.
GPT-4V has a clear structure and provides the longest answers among all models.
When evaluating low aesthetic quality images, the model scores are lower, which also indicates an insight that accurately identifying the shortcomings of an image is more challenging than praising one.
\oursmodel can still pinpoint the shortcomings of the images accurately, such as cluttered composition and overexposure, while other models tend to praise the images instead.
Using IDCP data could lead to better aesthetic discernment and expression ability.
\setlength{\abovecaptionskip}{2pt}

\setlength{\abovecaptionskip}{2pt} 
\section{Limitations}
\subsection{\oursmodel Limitations}
Despite the enhanced aesthetic abilities, the performance on general-purpose tasks, especially those that require heavy reasoning abilities and language tasks, has decreased. 
Hence, if \oursmodel is used for tasks other than aesthetic perception and understanding, it may generate undesired outcomes. 
Secondly, despite the improved accuracy, \oursmodel still performs worse than the  human. Therefore, they may not be able to replace humans directly. 
Thirdly, the Converted dataset from IDCP mostly comprises natural in-the-wild images.
The performance can be further refined by adding more dataset sources, including natural image datasets, artistic image datasets, AI Generated Content datasets.
Future work will focus on universal visual aesthetics, including aesthetic evaluation of videos.

\subsection{\oursbench Limitations}
Although we have already aimed to maintain a balance in the number of dimensions for each aesthetic attribute and the balance between options, there inevitably exists some numerical deviation in the collection process. We will continue to refine the benchmark.
Regarding the \textbf{description} task, we recognize that determining whether an MLLM answer aligns with the gold description is a subjective process that lacks an absolute standard. 
Human scores on the MLLM descriptions are also subject to individual differences. Despite employing the 5-round GPT-assisted evaluation protocol, which is currently the most trustworthy and reproducible approach, there may still be hallucinations (from GPT) that cannot be avoided. 
Moving forward, we will keep exploring better evaluation protocols for the low-level visual description task.
We will continue to expand the scope of the aesthetic evaluation system in \oursbench, including but not limited to art paintings, AIGC, videos, movie clips, and other samples along with their corresponding aesthetic dimensions.

\end{document}